# A Robust Deep Networks based Multi-Object Multi-Camera Tracking System for City Scale Traffic

Muhammad Imran Zaman[1], Usama Ijaz Bajwa[1]*, Gulshan Saleem[1] and Rana Hammad Raza[2]

[1]*Department of Computer Science, COMSATS University Islamabad, Lahore Campus, Lahore, Pakistan.
[2]Pakistan Navy Engineering College, National University of Sciences and Technology (NUST), Pakistan.

*Corresponding author(s). E-mail(s): usamabajwa@cuilahore.edu.pk;
Contributing authors: imranzaman.ml@gmail.com; gulshnsaleem26@gmail.com;
hammad@pnec.nust.edu.pk.

## Abstract

Vision sensors are becoming more important in Intelligent Transportation Systems (ITS) for traffic monitoring, management, and optimization as the number of network cameras continues to rise. However, manual object tracking and matching across multiple non-overlapping cameras pose significant challenges in city-scale urban traffic scenarios. These challenges include handling diverse vehicle attributes, occlusions, illumination variations, shadows, and varying video resolutions. To address these issues, we propose an efficient and cost-effective deep learning-based framework for Multi-Object Multi-Camera Tracking (MO-MCT). The proposed framework utilizes Mask R-CNN for object detection and employs Non-Maximum Suppression (NMS) to select target objects from overlapping detections. Transfer learning is employed for re-identification, enabling the association and generation of vehicle tracklets across multiple cameras. Moreover, we leverage appropriate loss functions and distance measures to handle occlusion, illumination, and shadow challenges. The final solution identification module performs feature extraction using ResNet-152 coupled with Deep SORT based vehicle tracking. The proposed framework is evaluated on the 5th AI City Challenge dataset (Track 3), comprising 46 camera feeds. Among these 46 camera streams, 40 are used for model training and validation, while the remaining six are utilized for model testing. The proposed framework achieves competitive performance with an IDF1 score of 0.8289, and precision and recall scores of 0.9026 and 0.8527 respectively, demonstrating its effectiveness in robust and accurate vehicle tracking.

## 1. Introduction

The rapid proliferation of sensors has led to the production of massive amounts of data, as well as the emergence of 5G technology and its growing acceptance. The fact that such data can now be analyzed at terminal devices makes it feasible to extract useful insights using the Internet of Things to raise operational efficiency and enhance overall outcome. This in turn makes it possible to improve the existing systems, which rely on manual inspection. Systems such as Intelligent Transportation Systems (ITS) are using Artificial Intelligence (AI) in order to improve traffic monitoring and optimization on a citywide scale [1]. These kinds of systems can be deployed to manage and monitor the traffic and assist autonomous vehicles.

ITS is soon becoming a need for the management of any city's transportation department as the number and diversity of vehicles in that city rapidly expand [1]. A huge cluster of network cameras with often non-overlapping fields of views is required to monitor and manage city traffic. Due to the enormous number of objects scattered across a volume of video feeds, manual object detection and tracking are impracticable. As a result, intelligent and autonomous vision-based solutions are required to intelligently detect, relate, and process objects and events of interest [2]. It also requires spatio-temporal correlation between identical objects acquired by non-overlapping cameras. In such scenarios, information pertaining to camera calibration is often unavailable and thus adding to the complexity of the problem. Different vehicle brands and models, color, size, pose, speed, vehicle-to-vehicle occlusions, backdrop clutter, illumination conditions, cast shadows, and video resolution all contribute to the complexity of the problem.

In order to address this problem, Multi-Object Multi-Camera Tracking (MO-MCT) systems are required. These systems take advantage of the most recent advancements in computer vision and are supported by development of deep neural networks [3, 4]. In this context, the terms Multi-Target Multi-Camera Tracking (MT-MCT) and Multi-Camera Multi-Object Tracking (MC-MOT) are interchangeably used for the same concept. When addressing city-scale urban traffic management, it is often necessary to perform object tracking and matching while simultaneously managing several sub-trajectories, also known as tracklets of the same vehicle across multiple non-overlapping cameras.

MO-MCT is an active research area with numerous applications, and many researchers have contributed significantly to this field [5-10]. Military-specific target detection and tracking, player tracking in a sporting event, crowd behavior analysis, activity recognition [2, 11, 12], and traffic monitoring and management systems [13, 14] are some of its applications. Object detection [15], object association [16, 17], and object tracking are the three fundamental functions of MO-MCT [18-20]. Object detection [21] is a key task that determines the location of an object in an image. In contrast, object association [22] provides correlation between detected items in subsequent frames. Finally, object tracking is used to assign IDs and labels to observed objects while keeping the label assignment consistent over frames. Detect-to-track and track-to-detect are two methodologies that the research community typically employs depending on the type of dataset/application. Mask R-CNN [23], YOLO-v3 [24], and Single Shot Multi-Box Detector (SSD12) [7] architecture-based object detection are used as baselines in this investigation. For object association, energy minimization using triplet loss and aggregation loss is used. Similarly, deep SORT [25], which draws its visual features from Mask R-CNN, is the method that is used for object tracking.

In the context of urban traffic, MO-MCT is connected with various issues, including blind spots [26] the assignment of labels to stationary items such as parked automobiles, and their association with dynamic objects. The distance and angle between the camera and the vehicle vary between several cameras. Furthermore, re-identification becomes difficult due to the objects appearing, disappearing, and reappearing in a single view. In any particular instance, non-overlapping camera viewpoints increase the complexity of the problem. In a similar manner, dynamic color corrections across cameras are required [27, 28] when the lighting conditions varies abruptly. In addition, the MO-MCT task is made more challenging by a number of parameters, including vehicle position and vehicle speed, both of which contribute to the appearance of motion blur.

The MO-MCT framework described in this study addresses these challenges and competes with existing approaches. In brief, the proposed study provides the following contributions:

- Introduced a novel four-step framework specifically designed for city-scale traffic monitoring, addressing the challenges of tracking multiple objects across multiple cameras. The framework focuses on achieving strong vehicle tracking and re-identification.

- Proposed an adaptive aggregation loss that accurately associates vehicle trajectories based on their respective weights, enabling correct trajectory association.
- Utilized inter-class non-maximum suppression in the proposed approach, effectively addressing the problem of overlapping objects in multi-camera tracking. This contribution significantly improves the accuracy and reliability of object tracking, particularly in dense traffic situations.
- Introduced a separate approach for deep visual feature generation using processed detections for Deep SORT, resulting in a significant reduction in training time for the multi-object multi-camera tracking system. This approach enhances system efficiency, especially in real-time applications, and contributes to the development of time-efficient, high-performance tracking systems.
- Enhanced vehicle re-identification across multiple cameras by employing ResNet-152, a powerful deep neural network architecture, and aggregate loss. This contribution significantly improves the accuracy and reliability of vehicle identification and tracking in traffic monitoring scenarios, enabling robust surveillance throughout the monitored area.

## 2. Literature Review

In the recent decade, extensive research on Single-Camera Multi-Object Tracking (SC-MOT) and Multi-Camera Multi-Object Tracking (MC-MOT) has been conducted (MC-MOT). MOT has been employed in a wide range of applications, including traffic monitoring and management, crowd behavior analysis, video surveillance, and human action recognition [11]. For the purpose of vehicle tracking, SC-MOT has used optical, semantic, and temporal features. It may also be regarded as a graph optimization problem [8], where the vertices of the graph represent identified objects and the edges reflect the similarity of two detections from separate frames. Object labelling and ID management become problematic for single-camera-based multi-object tracking due to intermittent multiple object entry-exit and occlusion issues in a single viewpoint traffic scenario.

In recent years, a number of different deep learning-based multi-object tracking algorithms that are designed to deal with MC-MOT have been reported. Although multi-object tracking using multiple video streams has the potential to dramatically increase object tracking across a wider region, it also introduces new challenges into the process. These are associated with a wide variety of camera configurations, each of which can be characterized by examining one of two primary scenarios. Using many cameras, for instance, allows for the recording of the same scene from a variety of views (i.e., partially to completely overlapping). As a consequence of this, it is challenging to correlate and interpret the information that is collected from various data points.

It's also possible that the target objects travel across many cameras, leading to the recording of various, non-overlapping scenes. In situations like these, object re-identification is of the utmost importance, and in order to carry out tracking, it is necessary to link the data from various cameras. There are many different designs for detecting objects, some of which are based on Residual Network [29], EfficientNet [30], the YOLO [31], VGG [32], and the Mask R-CNN [23]. These architectural choices include compromises in terms of performance, the nature of the dataset, and the context of the application.

Labbe et al. [33] introduced a multi-view multi-object pose estimate model based on EfficientNet [30] that can recover the 6D pose of numerous known objects. They tested their proposed technique on two challenging datasets: YCV-Video and T-LESS. The paper employs MT-MCT to recognize the various viewpoints of each vehicle captured by various camera streams.

Sun et al. [34] proposed a deep affinity network for multi-object tracking that uses the VGG architecture as a backbone for object detection. They used UA-DETRAC [35], MOT15 [36], and MOT17 [36] object tracking datasets in their study. The affinity estimator and deep track association in their proposed framework aid in the tracking of multiple objects in varying lighting conditions. Similarly, Tan et al. [31] presented a multi-target tracking algorithm based on the YOLO architecture for fast object detection.

Sequel to object detection, LSTM [37] is applied to find matching frames using pre-processed features. Euclidean distance is typically applied as a measure of association's degree of similarity. Peri et al. [38] proposed using Mask R-CNN-based detections to re-identify vehicles and track objects using several cameras. They used a re-identification method that involved computing the distance matrix of each vehicle's trajectory. They also applied agglomerative clustering to the trajectories of single camera tracking that resulted in multi-camera multi-object tracking.

By employing techniques like temporal and appearance features when using the detection-by-tracking paradigm to address object tracking difficulties.

While using detection-by-tracking paradigm, temporal and appearance features are used by various researchers to address object tracking challenges. The size, mobility, and placement of objects in temporal features are utilized in the bounding boxes. For instance, Tan et al. [31] presented multi-camera vehicle tracking based on visual and temporal features such as global images, regions, and tracking data from several locations. They used trajectory and calibration features to calculate the similarity of two objects in order to identify the cluster of each object for object re-identification and tracking.

Zhang et al. [5] proposed an online and real-time multi-camera object tracking system employing temporal information such as camera locations and time intervals, based on multiple offline methods provided for MT-MCT by various researchers. The authors of [39] have demonstrated how attribute semantic parsing for MT-MCT can be used to monitor automobiles in a reliable manner. The detections were linked using graph clustering and by employing local tracklets. In addition, they presented a spatiotemporal attention strategy for generating robust representations. A CNN-based model is used for object tracking and re-identification when appearance-based features [40, 41] are used. Similarly, the authors of [42] used CNN and k-means for multi-target multi-camera tracking and re-identification. They have also proposed evaluation measures that compare the performance of the tracking results. Adaptive weighted triplet loss for training and a new hard-identity mining technique are among their contributions.

Voigtlaender et al. [43] enhanced the Mask R-CNN to Track R-CNN by including the Euclidean distance matrix between each detection's association vectors. This aided in the association of objects and the assignment of tracking IDs. They also proposed the soft Multi-Object Tracking and Segmentation Accuracy (sMOTSA) metric to evaluate object tracking performance. To provide sophisticated pixel-level annotations for two separate tracking datasets, a semi-automatic annotation technique is used. They performed Track R-CNN with several association mechanisms (association head, mask intersection of union, Bbox center) and compared the results on the KITI MOTS dataset.

Similarly, authors of [44] proposed a motion assessment network and an appearance evaluation network that extract long-term properties to perform tracklet association. They used the MOT17 benchmark dataset and developed a new method for extracting tracklet features in order to perform better object association (vehicles and people). The authors of [45] worked on the re-identification problem and presented a Siamese architecture for extracting semantics from data at various depth levels. Using identity and similarity learning, this architecture integrated the data to accomplish a re-identification task. The researchers developed a striped pyramidal pooling block (SPPB) to extract multi-level information from image data in the shape of a pyramid. Identity and similarity are computed using an attention-based feature vector, which are then analyzed collectively in a mixed learning method. Three different Re-ID datasets are used to assess the proposed architecture (i.e. CUHK03 [46], MARKET1501 [47], and DUKE MTMC [48]).

The authors of [49] presented a real-time Fast-Constrained Dominant Set Clustering (FCDSC) approach. To construct coherent and compact clusters, FCDSC employs a set of constraints and a graph. A similarity score is produced using extracted features, which is then used to assign labels/IDs. Two benchmark datasets MOT challenge and Duke MTMC are used to train and validate their method.

It is evident from the existing literature that different researchers have attempted to address the MT-MCT problem, but there are still areas that require attention. Among these include cost-effective model design, addressing the issue of non-uniform illumination [50, 51], occlusion [52], and object association. To overcome these issues, we proposed an end-to-end framework built on spatiotemporal and trajectory-based features. The proposed study used vehicle trajectory weights to propose an adaptive aggregation loss for accurately associating and identifying them. The cost-effectiveness of the model is addressed by employing tracklets from adjacent cameras rather than all processing streams and performing data point merging to make the model robust to different variations.

## 3. Proposed Approach

The proposed multi-object multi-camera tracking framework (MO-MCT) accomplishes the objective in four stages, as shown in Figure 1. The first stage improves vehicle identification by using Non-Maximum Suppression (NMS) on Mask R-CNN [23] detections. The next stage performs vehicle re-identification by generating vehicle tracklets from multiple cameras using transfer learning. The fourth and final stage involves assigning tracking IDs based on the results of vehicle re-identification. To achieve multi-camera multi-object tracking, unique tracking IDs are synchronized using predefined criteria. The following sections describe the various stages of the development of the proposed approach.

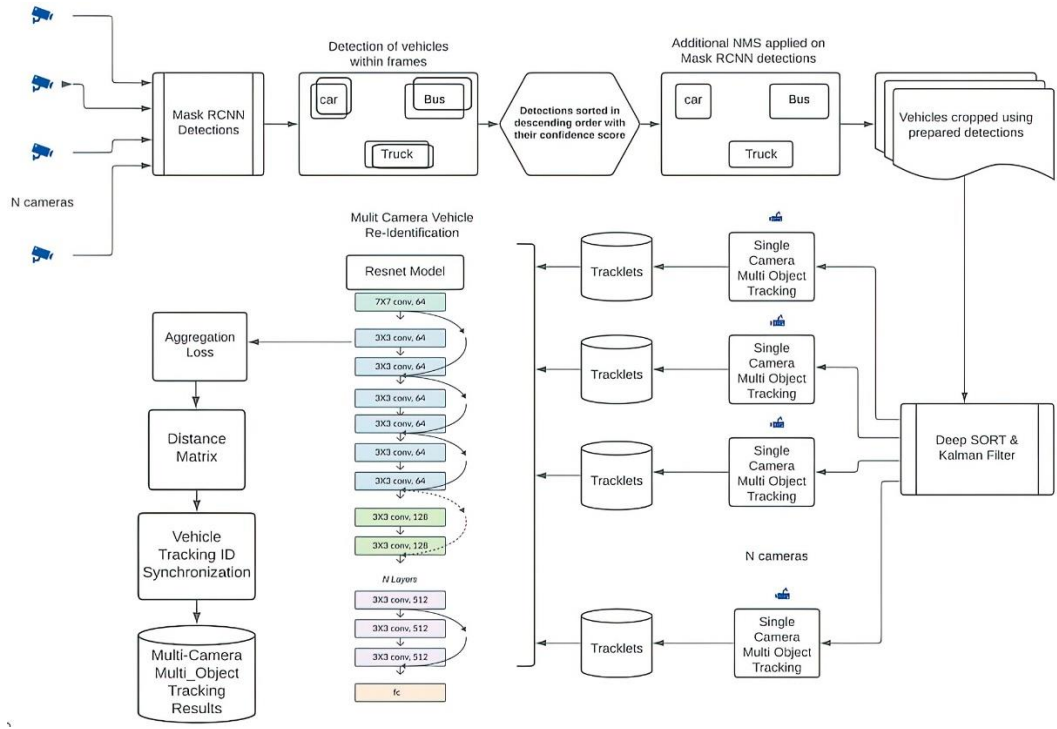

**Fig. 1** Illustration of proposed MO-MCT Framework

### 3.1. Dataset Description

The AI City Challenge dataset [53] is a well-known dataset containing multi-camera streams taken from 46 cameras over the course of 215.03 minutes. These 46 camera streams were captured from

16 crossroads in a mid-sized city in the United States. Except for one of the cameras (Camera No. 15), the resolution of the camera stream is 960p and 10 frames per second. The maximum distance between the two cameras that are the farthest apart is 4 kilometres. The video streams of the dataset cover highway variations, location types, intersections, and roadways. The dataset is divided into six scenarios whereas the Fig. 2 provide the depiction of camera locations on a map. Three scenarios are supplied for model training whereas two scenarios are supplied for validation; and one scenario is reserved for model testing. There are approximately 313,931 bounding boxes and 880 unique annotated vehicle identities. Annotations are added to automobiles that have been captured by at least two separate cameras. Furthermore, in each scenario, the offset from the start time for each video is provided, which is used for synchronization.

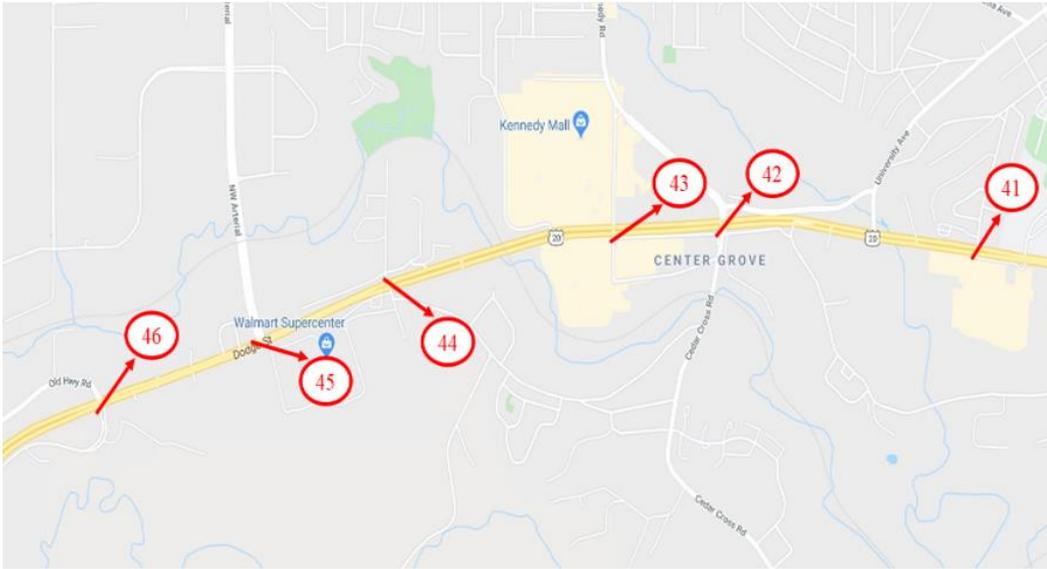

**Fig. 2** Depiction of camera locations in 5th AI City Challenge dataset on a map

The proposed approach uses Mask R-CNN detection as the first step followed by application of non-maximum suppression based on assumption that there are N camera streams. Data preparation, development of tracklets, and representation of vehicles through the use of a tracking algorithm are performed after that. In the end, vehicle re-identification through the use of the backbone model (i.e., ResNet-152) is performed resulting in multi-camera multi-object tracking.

## 3.2. Dataset Preparation and Object Detection

The tracking-by-detection paradigm was applied in the proposed approach. The first stage performs feature extraction for vehicle detection using a spatiotemporal approach. Fig. 1 depicts the layout of the MO-MCT framework, which begins with the application of Mask R-CNN based detections on each frame of video. The "instance segmentation network" is incorporated into the proposed MO-MCT framework. Mask R-CNN is a frame level object detector which additionally provide. It uses a CNN-based model as the foundation for feature extraction, such as ResNet101 [3] with a "feature pyramid network" [54]. The Region of Interest (RoI) and the region proposal network are represented by the feature representations. For each detection, multiple output heads are used to determine the object class, bounding box segmentation mask, and confidence score. To protect privacy, the AI City Challenge 2021 Track 3 dataset does not include vehicle license plate information.

In a traffic scenario, the target objects are vehicles such as buses, trucks, and cars. In most cases, the standard Mask R-CNN performs NMS [55]. Multiple vehicles may be classified as belonging to the same class in the case of multi-vehicle tracking and ID assignment, such as a car and a truck being classified as pickup truck. As a result, to achieve more precise vehicle detection, an additional interclass NMS is applied. Following the use of NMS, detected vehicles are sorted in descending order based on their confidence score. Sorted vehicles are selected one at a time and dropped if they do not fall below the 0.3 threshold representing the Intersection over the Union (IoU). Finally, these prepared vehicle detections are incorporated into the development and depiction of tracklets. As illustrated in Figure 3, the dataset is first pre-processed, which involves cropping of prepared detections, which are then saved as vehicle images in galleries. This pre-processed dataset is used to train the vehicle association and re-identification backbone model.

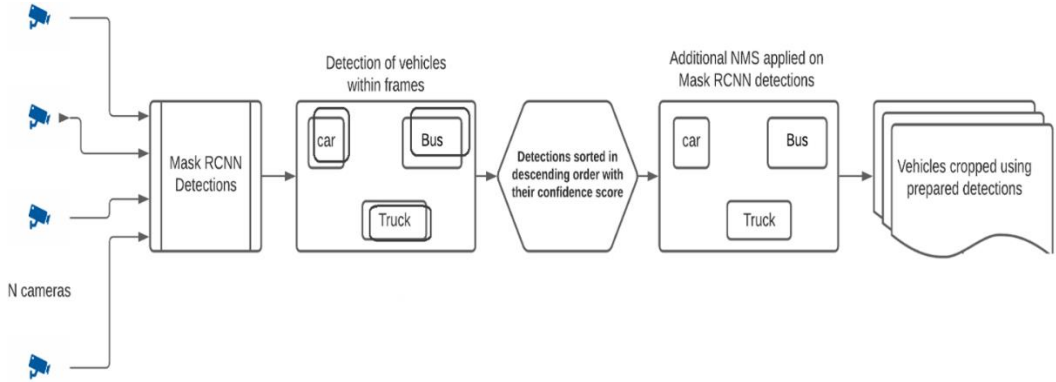

**Fig. 3** Data preparation and vehicle detection workflow

The labelling information is provided in the form of a text file providing the coordinates of the left and top lines, as well as the width and height of the vehicle in the stream. Based on the annotation, the dataset is separated into two parts: training and validation. Based on the given scenarios, 32 camera streams are used for training and 8 camera streams are used for validation. The validation dataset has 145 vehicles, whereas the training dataset contains 521 vehicles. Using test data from cameras 41 to 46, the S06 scenario is used to evaluate the model's performance for MO-MCT. Cropped images are stored in three different galleries: training, validation, and query. We select one query image to match with the training and validation sets in order to establish a unique tracking ID for each vehicle. There are the following number of cropped automobiles in each gallery folder:

- Number of images for training set = 206059
- Number of images for validation set = 24494
- Number of query images = 3028

### 3.3. Robust Tracklet Generation and Representation

The object tracking model and frame-level detection are merged together in order to produce tracklets. Deep SORT and MORT, two cutting-edge multi-target tracking models, are used in the proposed MO-MCT framework to correlate detections with tracklets [56]. Deep SORT algorithm combines constant velocity model and Kalman filter to predict object speed and location. The Kalman filter predicts the present state based on prior predictions and available detection. Deep SORT, on the other hand, uses deep visual elements as a criterion for association. Data-driven visual characteristics are considered to be more expressive than bounding boxes. MORT is yet another multi-object tracker that uses identified object properties to build single-camera object tracking. We reused the region of interest characteristics of the vehicle for Deep SORT and MORT from the Mask R-CNN to conduct the association criterion to reduce computational complexity. The association of fresh detection and existing tracklets is based on feature similarity and compliance to "spatial constraints." To construct the tracklets for each vehicle, detections are compared with previously confirmed tracklets based on their resemblance, and if a match is not spatially adjacent, it is discarded. Secondly, leftover detections are matched to previously verified tracklets using the bounding box IoU ("Intersection Over Union"). Finally, all unconfirmed tracklets are incorporated into matching and treated as new tracklets. In order to do vehicle re-identification using several cameras, we made use of the generated tracklets from each camera.

## 3.4. Vehicle Re-Identification

The backbone model receives its input in the form of a tracklets batch that contains cropped images taken from the training set $T_B \times W \times H \times 3$, as shown in Fig. 4. In this case, B represents the batch of tracklets that we constructed in the previous section, H represents the image's height, and W represents the image's width. We employed transfer learning for feature extraction and used the prepared dataset and generated tracklets to train the Residual Networks and VGG models. Aggregation loss ($loss_{agg}$) is utilised in backbone models for matching vehicles, and it is derived using the hard mining aggregates "cross-entropy loss" $loss_{xe}$ and "triplet loss" $loss_{tr}$. The margin loss is used to normalise and adjust the weight between hard positive loss $loss_{hp}$ and hard negative loss $loss_{hn}$. The hard negative is a vehicle image that looks a lot like other vehicle images. In contrast, a hard positive is described as a vehicle image that is distinct from another vehicle image. As a result, it is assumed that if the model can classify hard negative and hard positive samples, it should also be able to classify other samples. This approach overcomes the different viewpoints of each vehicle in order to appropriately classify it.

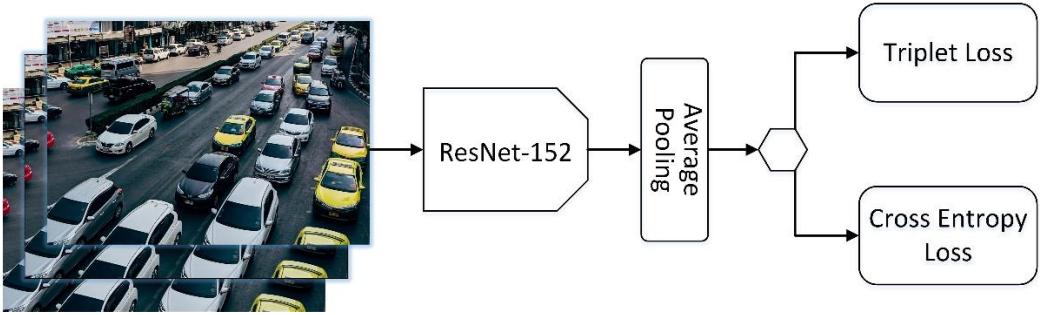

**Fig. 4** Vehicle Re-Identification training through backbone model using aggregation loss

The cross-triplet loss is represented by equations 1, 2 and 3.

$$loss_{tr} = max(0, loss_{hp} + loss_{hn} + \lambda) \quad (1)$$

$$loss_{hp} = \sum_{i=1}^{V} \sum_{j=1}^{B_i} max_{k=1}^{B_i} . max^{B_i} k = 1 \left( D \left( N_{feat}\left(I_j^i\right), N_f eat\left(I_p^q\right) \right) \right) \quad (2)$$

$$\sum_{i=1}^{V} \sum_{j=1}^{B_i} max_{p=1, q=1, p \neq q}{}^{B_i} . max^{B_i} k = 1 \left( D \left( N_{feat}\left(I_j^i\right), N_f eat\left(I_p^q\right) \right) \right) \quad (3)$$

$B_i$ represents the image of Vehicle $i$, $I_j^i$ represents the $j_{th}$ image of $I$ vehicle, and $D$ represents the distance function. whereas a hard negative loss is denoted by $loss_{hn}$ and a hard positive loss by $loss_{hp}$. According to the loss function losstr, positive distance should be minimal, while hard negative distance should be larger. In order to determine the aggregation loss, the cross-entropy loss and the triplet loss are combined.

Figure 5 depicts vehicle re-identification pipeline using a backbone model that computes the distance matrix of each vehicle with other vehicles frame by frame using the aggregation loss.

First, a batch of tracklets from a multi-object tracking model is fed into the algorithm. The given batch is then divided into two parts: a $Q_{\times N \times W \times H \times 3}$ query set with $N$ frames and a $G_{\times M \times W \times H \times 3}$ gallery set with $M$ frames. To extract vehicle attributes from both sets, we employed a CNN-based pre-trained backbone model. The distance between the extracted features of set $Q$ and the extracted features of set $G$ is finally calculated to generate vehicle tracklets for the MO-MCT results.

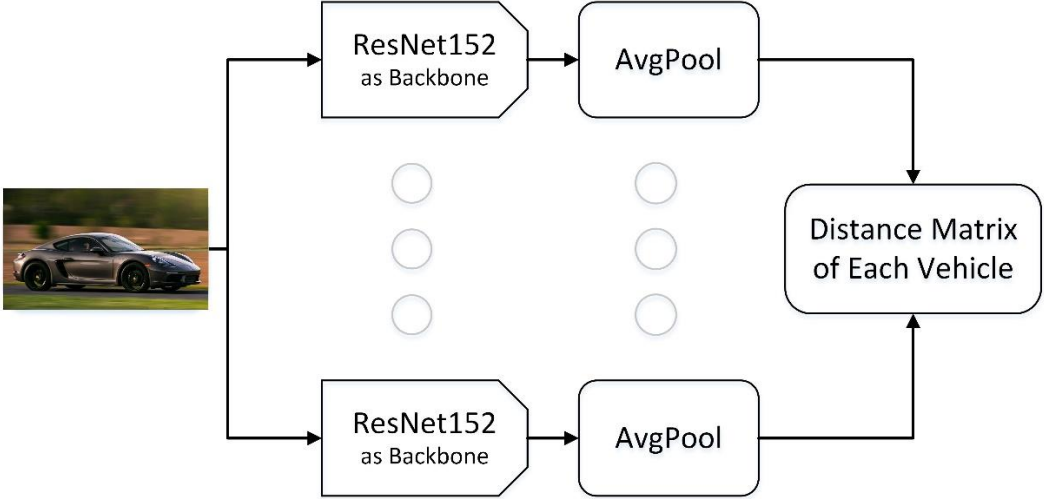

**Fig. 5** Depiction of vehicles Re-Identification pipeline

### 3.5. Tracklet ID Assignment

The proposed MO-MCT framework has the ability to accelerate the process that produces synchronized and reliable results. In order to keep the tracking results of previous multi-object tracking up to date, the distance matrix is used in accordance with the following rules:

- Sorting the tracklets by camera ID and comparing them to tracklets from adjacent cameras.
- Using the maximum threshold, removing tracklets with a large distance between query tracklets and all other gallery tracklets.
- If gallery tracklets from one camera match gallery tracklets from another camera and the distance between them is less than the synchronization threshold, tracking IDs are updated.

The minimum criterion for synchronization is set at 3, while the maximum threshold is set at 90. Because the target vehicle can travel to the left or right before appearing in a camera located further away, the tracklets of a particular camera view are not compared to the tracklets of all of the other camera views. According to the estimates, it will take many minutes for a car to arrive to Main Street no matter which way it turns, and the length of each video in the collection is only a few minutes. As a consequence of this, only the tracklets of two cameras that are next to one another are compared, and the distance between tracklets is determined in a sequential manner based on the camera ID.

## 4. Results & Discussion

### 4.1. Experimental Setup

Table 1 details the proposed MO-MCT framework's hyperparameters as well as the resource specifications used during experimentation. As an optimizer function, MO-MCT employs RMSProp with aggregate loss function, which was formed by combining the triplet loss and the entropy loss. All of the proposed MOC-MCOT frameworks' backbone models use aggregate loss whereas learning rate and weight decay were finalised after a few iterations. We experimented with different batch sizes and discovered that 128 is the best combination of model performance and computational resources. A larger batch size indicates that more memory is required on both system and GPU.

Table 1 System resources and hyperparameters

| Hyper Parameters | | Resources | |
|---|---|---|---|
| Batch Size | 128 | GPU | GTX 1080Ti |
| Optimizer | RMSProp | System Memory | 32 GB |
| Learning Rate | 0.001 | Disk Space (SSD) | 150 GB |
| Loss Function | Aggregation loss | Operating System | Linux |

## 4.2. Validation Results

In the validation set, video streams from eight cameras are used, while 32 cameras are used in the training set. Because the camera streams are split, every fifth stream is used for validation. We created tracklets using the MORT algorithm, and some of them had higher recall than others, although IDF1 was poor. Deep SORT, on the other hand, helped the proposed framework achieve good performance. On the other hand, when using Deep SORT, the proposed framework worked quite well. Figure 7 shows the training and validation accuracy, as well as the loss, during 200 epochs of final model training. The final model trained using ResNet152 has a terminal validation accuracy of 85.39 percent and a validation loss of 0.3850.

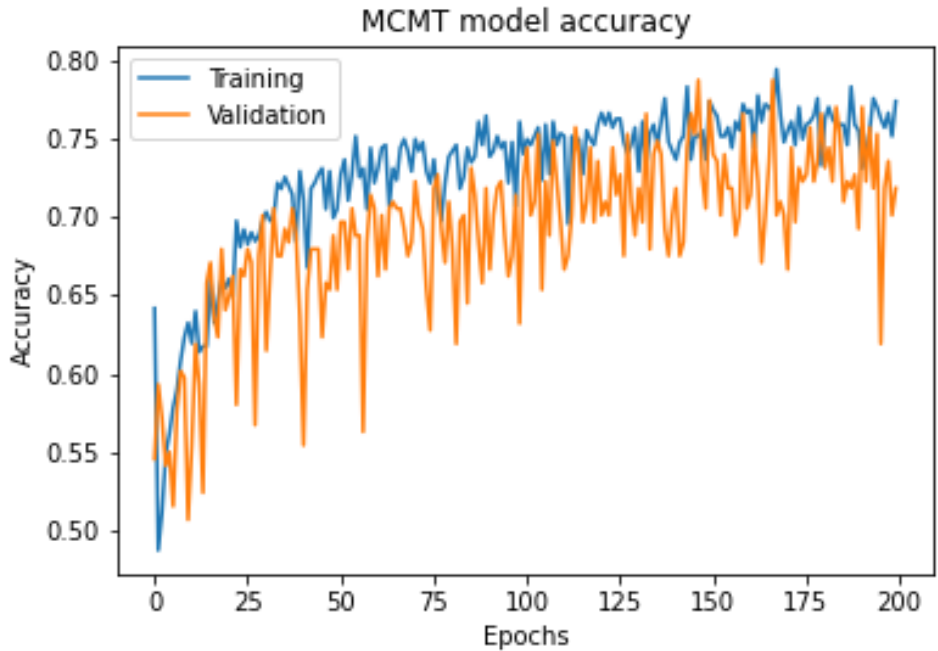

Fig. 6 Training and validation accuracy over 200 epochs.

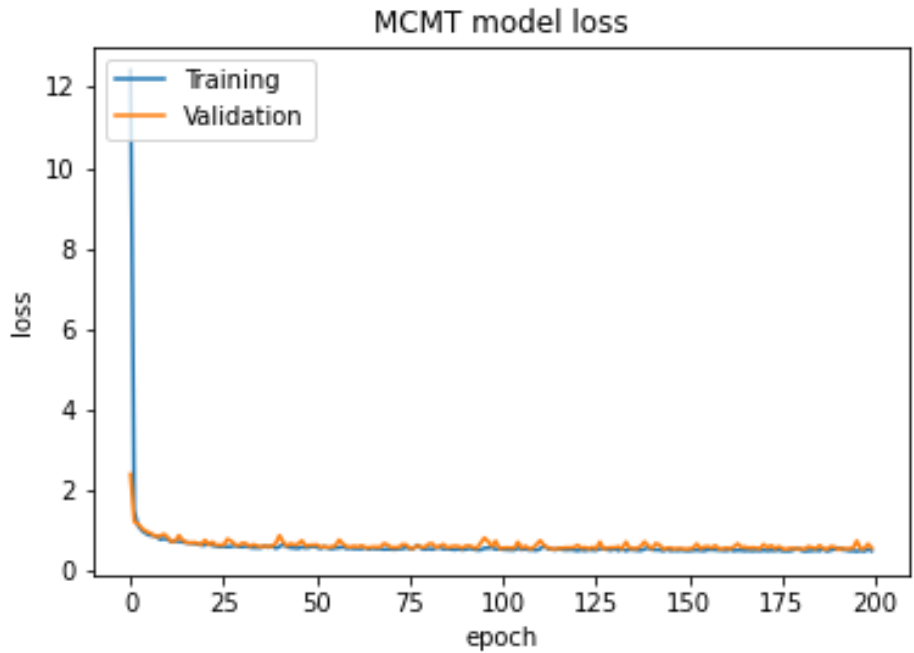

**Fig. 7** Training and validation accuracy and loss over 200 epochs.

## 4.3. Ablation Analysis

Extensive ablation analysis was conducted to determine the optimal network architecture and hyperparameter configuration for the deep learning architecture. The analysis involved evaluating various backbone architectures and their corresponding configurations. Tables 2 and 3 present the results of these ablation experiments, demonstrating the validation performance achieved by different network configurations.

The selection of the most suitable network architecture and hyperparameters was an iterative process. The loss function, batch size, learning rate, and other parameters were carefully chosen to achieve the optimal performance of the network. The ablation experiments demonstrated the efficacy of the designed approach, which is based on an optimal framework supported by the chosen backbone network architecture. Evaluation metrics such as IDF1 (the harmonic mean of precision and recall) were employed to compare the performance of different backbone models. Precision and recall were also utilized to assess the model's performance on the validation dataset. Four distinct backbone models were evaluated, each yielding unique outcomes as depicted in Table 2. Moreover, the tracklets were generated using the MORT and Deep SORT algorithms, and the results of these experiments are summarized in Table 2.

The ablation experiments revealed the performance of the proposed framework with different backbone models and tracking algorithms. The Deep SORT tracking algorithm-based variants provide consistently high performance in comparison to MORT based tracking algorithm. Similarly, ResNet based variants are superior in comparison to Vgg16 architecture and their performance correlate with the increasing complexity of their architecture. The reported results demonstrate that the approach utilizing ResNet-152 with the Deep SORT algorithm achieves the highest IDF1 score (0.8289), along with notable precision and recall values, showcasing its efficacy in vehicle tracking and re-identification within the traffic monitoring scenario.

Table 2 Experimental results on validation set (IDF1, Precision & Recall)

| Backbone Model | Trk. Algo. | IDF1 | Precision | Recall |
|---|---|---|---|---|
| Proposed Framework+ResNet152 | Mort | 0.6556 | 0.7211 | 0.7034 |
| Proposed Framework+ResNet101 | Mort | 0.6219 | 0.6935 | 0.7337 |
| Proposed Framework+ResNet50 | Mort | 0.6055 | 0.6398 | 0.6201 |
| Proposed Framework+VGG16 | Mort | 0.5689 | 0.5903 | 0.5816 |
| Proposed Framework+ResNet152 | Deep SORT | 0.8289 | 0.9026 | 0.8527 |
| Proposed Framework+ResNet101 | Deep SORT | 0.7974 | 0.8719 | 0.83367 |
| Proposed Framework+ResNet50 | Deep SORT | 0.7515 | 0.7933 | 0.7787 |
| Proposed Framework+VGG16 | Deep SORT | 0.6516 | 0.6977 | 0.6836 |

## 4.4. Testing Results

The trained models are tested on the testing set, and the model performance is evaluated using the S06 scenario. The IDF1 scores of each of the various backbone models are provided in Table 3.

Table 3 Experimental results on test set (IDF1, Precision & Recall)

| Backbone Network | IDF1 |
|---|---|
| Proposed Framework+ResNet 152 | 0.5126 |
| Proposed Framework+ResNet 101 | 0.5108 |
| Proposed Framework+ResNet 50 | 0.5048 |
| Proposed Framework+VGG16 | 0.4680 |

In terms of IDF1 scores, the proposed framework with ResNet152 provided optimal performance and outperformed other approaches. Cropping and data augmentation are employed in the training process, and an aggregation loss function is used as an optimal design decision. ResNet152 learns a more extensive set of features than Deep SORT does whereas Deep SORT contains reused visual features. On the test data, the IDF1 score obtained by ResNet152 is 0.2526, which is the highest achieved score. The IDF1 scores obtained by ResNet50, ResNet101, and VGG16 are respectively 0.2508, 0.2448, and 0.2080. In the 5th AI City Challenge 2021, the completed model was entered with an IDF1 score of 0.2526 and a team ID of 60.

## 4.5. Comparison with Existing Approaches

The comparison of best performing model (Proposed Framework+ResNet 152) is performed with eight existing methods. The reported validation results obtained using the validation dataset are compared using IDF1 score which provides a balanced representation of precision and recall reported by the model. The proposed approach has provided highest percentage IDF1 score in comparison to eight existing methods. The results of the comparison are reported in Table 4

Table 4 Comparison of the proposed approach with existing methods

| # | Method | IDF1 Score |
|---|---|---|
| 1 | Alibaba-UCAS | 80.95% |
| 2 | Baidu | 77.87% |
| 3 | SJTU | 76.51% |
| 4 | Fraunhofer | 69.10% |

| # | Method | IDF1 Score |
|---|--------|------------|
| 5 | Fiberhome | 57.63% |
| 6 | NTU | 54.58% |
| 7 | KAIST | 54.52% |
| 8 | NCCU-NYMCTU-UA | 13.43% |
| 9 | Proposed | 82.89% |

The strength of our proposed approach comes from the optimal design decision. Due to feature reuse and the use of tracklets from adjacent cameras rather than processing all video streams, the suggested solution is cost-effective. Strong multi-camera tracking IDs are achievable through the use of a combination of information obtained from multiple data points, such as adjacent cameras. This makes it possible to perform simultaneous multi-camera tracking. Figure 8 depicts a randomly chosen scenario in which vehicle images were recorded by three separate cameras and the suggested method still performed the detection and tracking accurately. Experiments have shown that the proposed technique can handle a reasonable level of vehicle occlusion and is robust to considerable variations in lighting conditions, look angles, and the existence of multiple similar-looking objects.

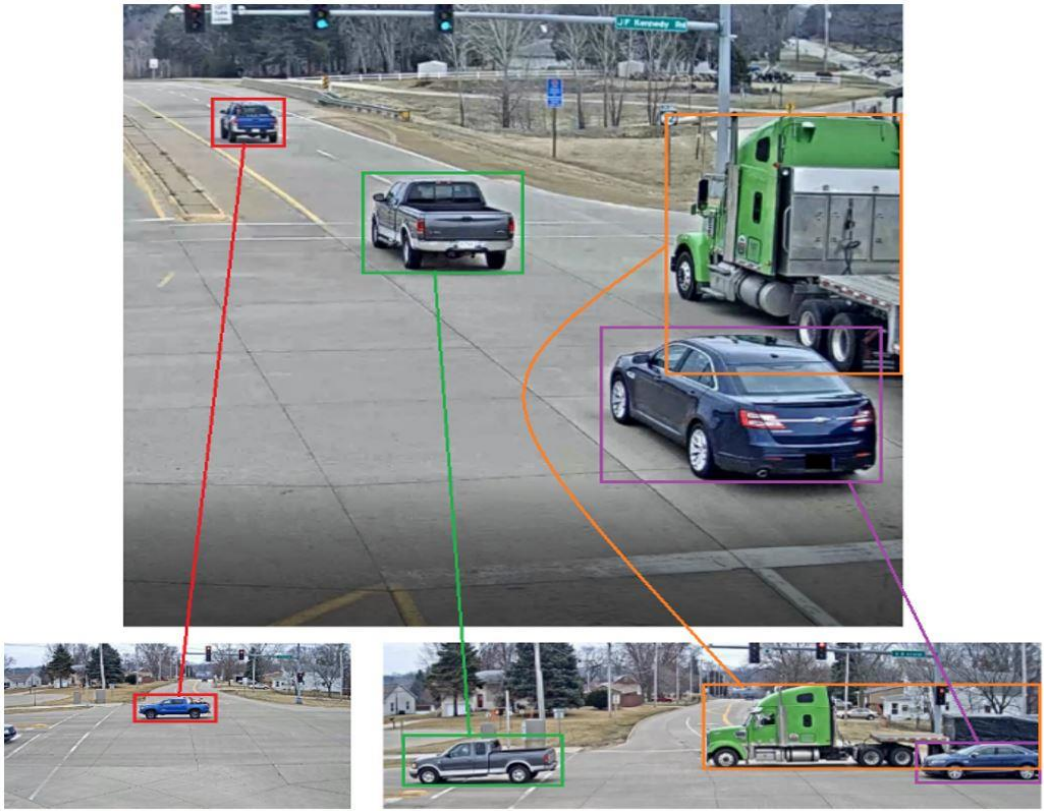

**Fig. 8** Demonstration of vehicles tracked by multiple cameras.

## 5. Conclusion

Due to the growing deployment of surveillance systems in urban and traffic monitoring, multi-object multi-camera tracking is becoming an increasingly important challenge. It is important to remember however, that there are small differences between traffic monitoring and crowd surveillance. The study focused on the problem of traffic monitoring and presented a four-step multi-object multi-camera tracking (MO-MCT) framework for city-scale traffic. The proposed framework is evaluated on the 5th AI City Challenge dataset (Track 3), which contains data from 46 cameras and presents challenging scenarios. The study's main objective is to achieve strong vehicle tracking using a multi-camera stream, as well as vehicle re-identification across several cameras. Due to the use of interclass non-maximum suppression, the proposed approach also overcomes the overlapping issue. Furthermore, rather than generating separate deep visual features for Deep SORT, our approach employs processed detections, resulting in a significant reduction in training time. Furthermore, ResNet-152 and aggregate loss are employed for re-identification, and instead of processing all video streams, tracklets for each vehicle are created from neighboring cameras for efficient MO-MCT. The proposed framework is less computationally intensive and eliminates the problem of overlap and occlusion. On the 5th AI City challenge dataset, the proposed solution achieved a score of 0.8289 for the IDF1 metric and scores of 0.9026 and 0.8527, respectively, for precision and recall.

## 6. Declarations

- **Funding:** This study acknowledges partial support from the National Center of Big Data and Cloud Computing (NCBC) and HEC of Pakistan for conducting this research.

- **Conflict of interest:** The authors declare that they have no known competing financial interests or personal relationships that could have appeared to influence the work reported in this paper.

- **Availability of Data and Materials:** The dataset for 5th AI City Challenge, Track-3 is used for these experiments and is available for use after registration at the link: https://www.aicitychallenge.org/2021-data-and-evaluation/

- **Code Availability:** The code and trained models can be obtained from the project repository: https://github.com/imranzaman5202/MO-MCT

- **Authors' Contributions:** Zaman: conception, implementation, writeup and revision; Bajwa: conception, supervision and revision; Saleem: implementation, writeup and revision; Raza conception, supervision and revision.

## References

[1] Lv, Z., R. Lou, and A.K. Singh, *AI empowered communication systems for intelligent transportation systems.* IEEE Transactions on Intelligent Transportation Systems, 2020. **22**(7): p. 4579-4587.

[2] Saleem, G., U.I. Bajwa, and R.H. Raza, *Toward human activity recognition: a survey.* Neural Computing and Applications, 2023. **35**(5): p. 4145-4182.

[3] He, K., et al. *Deep residual learning for image recognition*. in *Proceedings of the IEEE conference on computer vision and pattern recognition*. 2016.

[4] Ahmed, N., et al., *Deep ensembling for perceptual image quality assessment.* Soft Computing, 2022. **26**(16): p. 7601-7622.

[5] Zhang, X. and E. Izquierdo. *Real-time multi-target multi-camera tracking with spatial-temporal information*. in *2019 IEEE Visual Communications and Image Processing (VCIP)*. 2019. IEEE.

[6] Wu, Y., J. Lim, and M.-H. Yang. *Online object tracking: A benchmark*. in *Proceedings of the IEEE conference on computer vision and pattern recognition*. 2013.

[7] Liu, W., et al. *Ssd: Single shot multibox detector*. in *Computer Vision–ECCV 2016: 14th European Conference, Amsterdam, The Netherlands, October 11–14, 2016, Proceedings, Part I 14*. 2016. Springer.

[8] Kumar, R., G. Charpiat, and M. Thonnat. *Multiple object tracking by efficient graph partitioning*. in *Computer Vision--ACCV 2014: 12th Asian Conference on Computer Vision, Singapore, Singapore, November 1-5, 2014, Revised Selected Papers, Part IV 12*. 2015. Springer.

[9] Tang, S., et al. *Subgraph decomposition for multi-target tracking*. in *Proceedings of the IEEE Conference on Computer Vision and Pattern Recognition*. 2015.

[10] Schofield, K. and N.R. Lynam, *Vehicle blind spot detection display system*. 1998, Google Patents.
[11] Saleem, G., et al., *Efficient anomaly recognition using surveillance videos.* PeerJ Computer Science, 2022. **8**: p. e1117.
[12] Saleem, G., U.I. Bajwa, and R.H. Raza, *Surveilia: Anomaly Identification Using Temporally Localized Surveillance Videos.* Available at SSRN 4308311.
[13] Saleem, M., et al., *Smart cities: Fusion-based intelligent traffic congestion control system for vehicular networks using machine learning techniques.* Egyptian Informatics Journal, 2022. **23**(3): p. 417-426.
[14] Yuan, Y., Z. Xiong, and Q. Wang, *VSSA-NET: Vertical spatial sequence attention network for traffic sign detection.* IEEE transactions on image processing, 2019. **28**(7): p. 3423-3434.
[15] Wang, Q., et al., *Hybrid feature aligned network for salient object detection in optical remote sensing imagery.* IEEE Transactions on Geoscience and Remote Sensing, 2022. **60**: p. 1-15.
[16] Ning, X., et al., *HCFNN: high-order coverage function neural network for image classification.* Pattern Recognition, 2022. **131**: p. 108873.
[17] Wang, C., et al., *Uncertainty estimation for stereo matching based on evidential deep learning.* Pattern Recognition, 2022. **124**: p. 108498.
[18] Li, P., et al. *Spatio-temporal Consistency and Hierarchical Matching for Multi-Target Multi-Camera Vehicle Tracking*. in *CVPR Workshops*. 2019.
[19] Kohl, P., et al. *The mta dataset for multi-target multi-camera pedestrian tracking by weighted distance aggregation*. in *Proceedings of the IEEE/CVF Conference on Computer Vision and Pattern Recognition Workshops*. 2020.
[20] Sharma, A., S. Anand, and S.K. Kaul, *Intelligent querying for target tracking in camera networks using deep q-learning with n-step bootstrapping.* Image and Vision Computing, 2020. **103**: p. 104022.
[21] Benali Amjoud, A. and M. Amrouch. *Convolutional neural networks backbones for object detection*. in *Image and Signal Processing: 9th International Conference, ICISP 2020, Marrakesh, Morocco, June 4–6, 2020, Proceedings 9*. 2020. Springer.
[22] Ma, C., et al. *Deep association: End-to-end graph-based learning for multiple object tracking with conv-graph neural network*. in *Proceedings of the 2019 on International Conference on Multimedia Retrieval*. 2019.
[23] He, K., et al. *Mask r-cnn*. in *Proceedings of the IEEE international conference on computer vision*. 2017.
[24] Qiu, Z., et al., *Vision-based moving obstacle detection and tracking in paddy field using improved yolov3 and deep SORT.* Sensors, 2020. **20**(15): p. 4082.
[25] Wang, Q., et al., *MTCNN-KCF-deepSORT: Driver Face Detection and Tracking Algorithm Based on Cascaded Kernel Correlation Filtering and Deep SORT*. 2020, SAE Technical Paper.
[26] Schofield, K. and N.R. Lynam, *Vehicle blind spot detection display system*. 1999, Google Patents.
[27] Ahmed, N., H.M.S. Asif, and H. Khalid, *PIQI: perceptual image quality index based on ensemble of Gaussian process regression.* Multimedia Tools and Applications, 2021. **80**(10): p. 15677-15700.
[28] Ahmed, N. and H.M.S. Asif. *Ensembling convolutional neural networks for perceptual image quality assessment*. in *2019 13th International Conference on Mathematics, Actuarial Science, Computer Science and Statistics (MACS)*. 2019. IEEE.
[29] Zhang, K., et al., *Residual networks of residual networks: Multilevel residual networks.* IEEE Transactions on Circuits and Systems for Video Technology, 2017. **28**(6): p. 1303-1314.
[30] Tan, M. and Q. Le. *Efficientnet: Rethinking model scaling for convolutional neural networks*. in *International conference on machine learning*. 2019. PMLR.
[31] Tan, L., et al. *A multiple object tracking algorithm based on YOLO detection*. in *2018 11th International Congress on Image and Signal Processing, BioMedical Engineering and Informatics (CISP-BMEI)*. 2018. IEEE.
[32] Simonyan, K. and A. Zisserman, *Very deep convolutional networks for large-scale image recognition.* arXiv preprint arXiv:1409.1556, 2014.
[33] Labbé, Y., et al. *Cosypose: Consistent multi-view multi-object 6d pose estimation*. in *Computer Vision–ECCV 2020: 16th European Conference, Glasgow, UK, August 23–28, 2020, Proceedings, Part XVII 16*. 2020. Springer.

[34] Sun, S., et al., *Deep affinity network for multiple object tracking.* IEEE transactions on pattern analysis and machine intelligence, 2019. **43**(1): p. 104-119.

[35] Wen, L., et al., *UA-DETRAC: A new benchmark and protocol for multi-object detection and tracking.* Computer Vision and Image Understanding, 2020. **193**: p. 102907.

[36] Yoon, K., et al., *Oneshotda: Online multi-object tracker with one-shot-learning-based data association.* IEEE Access, 2020. **8**: p. 38060-38072.

[37] Kulkarni, P., et al. *Key-track: A lightweight scalable lstm-based pedestrian tracker for surveillance systems*. in *Image Analysis and Recognition: 16th International Conference, ICIAR 2019, Waterloo, ON, Canada, August 27–29, 2019, Proceedings, Part II 16*. 2019. Springer.

[38] Peri, N., et al. *Towards real-time systems for vehicle re-identification, multi-camera tracking, and anomaly detection*. in *Proceedings of the IEEE/CVF Conference on Computer Vision and Pattern Recognition Workshops*. 2020.

[39] He, Y., et al. *City-scale multi-camera vehicle tracking by semantic attribute parsing and cross-camera tracklet matching*. in *Proceedings of the IEEE/CVF Conference on Computer Vision and Pattern Recognition Workshops*. 2020.

[40] Wang, C., et al., *Learning discriminative features by covering local geometric space for point cloud analysis.* IEEE Transactions on Geoscience and Remote Sensing, 2022. **60**: p. 1-15.

[41] Ahmed, N. and H.M.S. Asif, *Perceptual Quality Assessment of Digital Images Using Deep Features.* Computing & Informatics, 2020. **39**(3).

[42] Ristani, E., et al. *Performance measures and a data set for multi-target, multi-camera tracking*. in *Computer Vision–ECCV 2016 Workshops: Amsterdam, The Netherlands, October 8-10 and 15-16, 2016, Proceedings, Part II*. 2016. Springer.

[43] Voigtlaender, P., et al. *Mots: Multi-object tracking and segmentation*. in *Proceedings of the ieee/cvf conference on computer vision and pattern recognition*. 2019.

[44] Zhang, Y., et al., *Long-term tracking with deep tracklet association.* IEEE Transactions on Image Processing, 2020. **29**: p. 6694-6706.

[45] Martinel, N., G.L. Foresti, and C. Micheloni, *Deep pyramidal pooling with attention for person re-identification.* IEEE Transactions on Image Processing, 2020. **29**: p. 7306-7316.

[46] Qian, X., et al. *Multi-scale deep learning architectures for person re-identification*. in *Proceedings of the IEEE international conference on computer vision*. 2017.

[47] Zheng, L., et al. *Scalable person re-identification: A benchmark*. in *Proceedings of the IEEE international conference on computer vision*. 2015.

[48] Gou, M., et al. *Dukemtmc4reid: A large-scale multi-camera person re-identification dataset*. in *Proceedings of the IEEE conference on computer vision and pattern recognition workshops*. 2017.

[49] Tesfaye, Y.T., et al., *Multi-target tracking in multiple non-overlapping cameras using fast-constrained dominant sets.* International Journal of Computer Vision, 2019. **127**: p. 1303-1320.

[50] Sun, H., et al. *Mvp matching: A maximum-value perfect matching for mining hard samples, with application to person re-identification*. in *Proceedings of the IEEE/CVF International Conference on Computer Vision*. 2019.

[51] Naphade, M., et al. *The 2019 AI City Challenge*. in *CVPR workshops*. 2019.

[52] Li, Y. and X. Wang, *Person Re-Identification Based on Joint Loss and Multiple Attention Mechanism.* Intelligent Automation & Soft Computing, 2021. **30**(2).

[53] Naphade, M., et al. *The 5th ai city challenge*. in *Proceedings of the IEEE/CVF Conference on Computer Vision and Pattern Recognition*. 2021.

[54] Kim, S.-W., et al. *Parallel feature pyramid network for object detection*. in *Proceedings of the European conference on computer vision (ECCV)*. 2018.

[55] Hosang, J., R. Benenson, and B. Schiele. *Learning non-maximum suppression*. in *Proceedings of the IEEE conference on computer vision and pattern recognition*. 2017.

[56] Wang, Z., et al. *Towards real-time multi-object tracking*. in *Computer Vision–ECCV 2020: 16th European Conference, Glasgow, UK, August 23–28, 2020, Proceedings, Part XI 16*. 2020. Springer.